\documentclass[sigconf,nonacm]{acmart}
\setcopyright{none}
\acmConference[WWW '27]{The ACM Web Conference 2027}{May 10--14, 2027}{Dublin, Ireland}
\acmYear{2027}
\usepackage{booktabs}
\usepackage{multirow}
\usepackage{pifont}
\usepackage{xspace}
\usepackage{amsmath}
\usepackage{graphicx}
\usepackage{subcaption}
\usepackage{algorithm}
\usepackage{algpseudocode}
\newcommand{\sys}{Surplus\xspace}
\newcommand{\unspent}{unspent error budget\xspace}

\newtheorem{proposition}{Proposition}

\begin{document}

\title{Semantic Caches for LLM Web Services Leave Their Error Budget Unspent}

\author{Hanshuai Cui}
\affiliation{\institution{School of Artificial Intelligence, Beijing Normal University, Beijing 100875, China}\city{}\country{}}
\affiliation{\institution{Institute of Artificial Intelligence and Future Networks, Beijing Normal University, Zhuhai 519087, China}\city{}\country{}}
\author{Qianli Ma}
\affiliation{\institution{Institute of Artificial Intelligence and Future Networks, Beijing Normal University, Zhuhai 519087, China}\city{}\country{}}
\author{Zhiqing Tang}
\affiliation{\institution{Institute of Artificial Intelligence and Future Networks, Beijing Normal University, Zhuhai 519087, China}\city{}\country{}}

\begin{abstract}
Web services built on large language models (LLMs) receive many paraphrases of earlier queries, which a semantic cache answers from stored responses.
A reused answer can be wrong, so wrong answers must stay below a share $\delta$ of the queries.
Existing caches reuse above a fixed similarity threshold or, like vCache, learn a per-entry rule that bounds each query's error by $\delta$.
However, fixed thresholds break the budget on new streams, and per-entry rules need labels on every entry before reusing it.
We find an \unspent, as vCache realizes only 12--55\% of $\delta$ on three public query streams.
\sys spends this budget where reuse is most likely correct.
It ranks queries by a risk score shared across paraphrase clusters and opens risk bins as far as an audited error estimate allows.
Free labels from every LLM call and a ledger of realized errors keep this threshold safe under drift.
On five streams, including ORCAS search queries, \sys holds $\delta$ in every test cell, which no baseline does.
On the shuffled streams it serves 40\% more hits than vCache, and its pooled hit rate beats the best baseline holding $\delta$ in each cell by 4.10 points.
In a vLLM testbed it cuts the mean latency by 31.59\% at the highest load, against 23.67\% for vCache.
\end{abstract}


\maketitle

\begin{figure}[t]
\centering
\includegraphics[width=\linewidth]{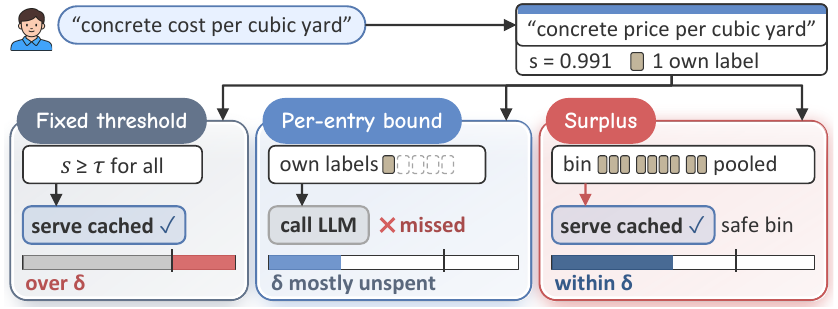}
\caption{\sys serves this Web search paraphrase within $\delta$, while a fixed threshold overspends $\delta$ and a per-entry bound calls the LLM. Its cached entry holds one label.}
\Description{Example schematic. The query concrete cost per cubic yard matches the cached query concrete price per cubic yard with similarity 0.991, and that entry holds one label of its own. Three lanes follow the query. A fixed threshold serves the cached answer, and its error bar runs past the budget. A per-entry bound calls the LLM because the entry has a single label, a missed reuse, and its error bar fills only part of the budget. Surplus serves the cached answer from a risk bin that pools labels across entries, and its error bar stays within the budget.}
\label{fig:teaser}
\end{figure}

\section{Introduction}
\label{sec:intro}

Web services now answer a growing share of user requests with large language models (LLMs), from answer boxes in Web search to chat assistants~\cite{gpt4o,qwen3,chatbotarena}.
These requests repeat.
Search engines have long cached results because a small set of queries recurs~\cite{silverstein,baezayates,fagni}. LLM services see the same questions again in new words, as in the 120,000 test queries we take from the ORCAS search log~\cite{orcas} through the SearchQueries benchmark~\cite{vcache}.
A semantic cache sits in front of the model as one layer of the Web serving infrastructure~\cite{gptcache,vllm} and returns a stored answer when a new query is close enough to a cached one, which saves a full LLM call per hit (Figure~\ref{fig:teaser}).
However, a paraphrase can ask something different, so every hit risks a wrong answer, and an operator needs to keep the share of queries that receive one within a budget $\delta$~\cite{vcache}.
A semantic cache is thus part of serving LLMs on the Web, a topic of the Web Infrastructure and Agentic Systems track.

Existing semantic caches decide a hit in three ways.
Caches with a fixed threshold reuse an answer whenever the embedding similarity exceeds a constant, as GPTCache~\cite{gptcache} and later systems do~\cite{gptsemcache,contextcache}.
Per-entry caches learn a rule for each cached entry.
vCache~\cite{vcache} fits a curve from similarity to correctness for every entry and explores with a probability that bounds each query's error by $\delta$, and MVR-cache~\cite{mvrcache} keeps this rule on a learned multi-vector similarity.
Global caches calibrate one threshold for all entries, either on a labeled prefix of the stream with conformal risk control~\cite{crc,gentleconf} or online~\cite{vcache}.

However, these designs spend the error budget poorly, because they overlook how unevenly risk is spread over the queries of a stream.
A fixed threshold ignores $\delta$ altogether, so it lands on either side of the budget.
A threshold calibrated on a prefix assumes that the prefix looks like the rest of the stream, which a growing cache breaks~\cite{beyondexch}, so it overspends once the stream drifts (Figure~\ref{fig:motiv}c).
A per-entry rule must collect labels on each entry before it reuses it and charges $\delta$ to every query separately, so it stays safe but sends reusable queries to the LLM (Figure~\ref{fig:motiv}a and b).
We measured these designs on three public query streams in shuffled order, LmArena chat prompts, Classification prompts and SearchQueries from Web search, and on two drifting streams, and identify an \emph{\unspent}.
On the shuffled streams 33--64\% of vCache's decisions meet an entry with fewer than four labels of its own, where it serves no hit (Figure~\ref{fig:measured}a).
Figure~\ref{fig:teaser} shows a fixed threshold and a per-entry bound on one paraphrase, with a threshold of 0.97 and bars that give the error of each design on SearchQueries against $\delta=1\%$.

Two obstacles keep this budget unspent.
\textbf{(C1)} \emph{How can a cache spend $\delta$ on the queries most likely to be reused correctly without overspending it?}
Per-entry estimates are sparse, since most entries have only a few labels, while one global threshold treats a near-duplicate and a loose paraphrase alike and cannot tell how much error a looser threshold would add.
\textbf{(C2)} \emph{How can it keep $\delta$ when the stream drifts?}
Labels come only from the queries the cache sends to the LLM, which are exactly the ones it did not serve, so an error estimate built at one point of the stream goes stale and an optimistic model overspends before any check notices.

We propose \sys, a semantic cache named for the error budget that per-entry caches leave unspent (Figure~\ref{fig:overview}).
First, to address C1, \sys scores each query by the risk of its nearest entry with a model shared across clusters of paraphrases. It opens risk bins in order of risk only as far as an audited estimate of the served error stays below the budget.
Second, for C2, every query that reaches the LLM labels its risk bin at no extra cost, and an error ledger compares the errors spent so far with the budget accrued so far and tightens the threshold as soon as they run ahead.
We validate \sys on these five streams against eight baselines and a hindsight oracle, and in a vLLM serving testbed.
It holds $\delta$ in every test cell, which no baseline does (\S\ref{sec:main}).
On the shuffled streams its pooled hit rate is 5.71 points above vCache's, which is 40\% more hits, and 4.10 points above the best baseline that holds $\delta$.
In serving it cuts the mean end-to-end latency at the highest load by 31.59\% where vCache cuts 23.67\% (\S\ref{sec:serving}).
In summary, we make three contributions.
\begin{itemize}
\item We identify an \unspent. Per-entry caches such as vCache, run at the target budget, realize only 12--55\% of it on public query streams (Appendix~\ref{app:cells}), because they lack labels and charge every query separately. Fixed thresholds miss it in either direction.
\item We design \sys, a semantic cache that ranks queries by a shared risk score and opens them under a calibrated risk threshold, which addresses C1, and keeps the threshold safe under drift with free labels and an error ledger that back each other up, which addresses C2.
\item Our evaluation spans search, chat and classification streams, including ORCAS search queries, with controls, capacity limits, judge and human reviews, generality checks and an end-to-end serving testbed.
\end{itemize}

\section{Related Work and Motivation}
\label{sec:related}

\subsection{Related Work}
\label{sec:rw}

\paragraph{Semantic caches for LLM services.}
A semantic cache reuses the answer of a similar earlier query.
GPTCache~\cite{gptcache} and later systems serve above a fixed similarity threshold, with user-side caches in MeanCache~\cite{meancache}, embedding ensembles~\cite{ensembleemb}, multi-turn context~\cite{contextcache} and engineering reports~\cite{gptsemcache,closeenough}.
SISO~\cite{siso} caches centroids with a dynamic threshold, and category-aware caching~\cite{categorycache} sets thresholds per query category, but neither bounds the error.
vCache~\cite{vcache} introduced a user-set error budget for semantic caches and learns a decision per entry. MVR-cache~\cite{mvrcache} keeps the vCache rule on a learned multi-vector similarity. Krites~\cite{krites} promotes answers that an asynchronous LLM judge approves but sets no error budget. FreshCache~\cite{freshcache} bounds the risk that a reused Web answer is stale rather than wrong.
Eviction and admission work learns which entries to keep~\cite{evictlearn,contsemcache,clever}, attacks and defenses target the similarity key~\cite{keycollision,lacache,deletiongain}, and operational studies tune the embedding model~\cite{opgap}.
\sys changes none of these parts.
It decides only when to serve, and we run it with other embeddings and under a least-recently-used (LRU) cap (\S\ref{sec:general}).

\paragraph{Risk control and calibration.}
Selective prediction abstains in order to bound the risk~\cite{selective}, and conformal prediction and risk control turn held-out labels into a threshold with a finite-sample guarantee~\cite{gentleconf,crc,ltt}, which holds when the data are exchangeable~\cite{beyondexch}.
Adaptive conformal inference~\cite{aci}, confidence sequences~\cite{confseq} and testing by betting~\cite{betting} track a risk online, and calibration fixes a model's scale~\cite{calibration}.
Our Conformal baseline represents this family with a split conformal threshold calibrated on a prefix of the stream.
\sys differs in two ways.
It calibrates a threshold on a ranked risk score with the labels that cache misses provide anyway, and it guards the realized error with a ledger instead of calibrating once.

\paragraph{Web caching and LLM serving.}
Web proxies and search engines cache documents and result pages under skewed query popularity~\cite{breslau,silverstein,baezayates,fagni}, and approximate caches reuse results for similar vector searches~\cite{aker} and diffusion prompts~\cite{nirvana}.
LLM serving systems batch, page and disaggregate inference~\cite{vllm,sglang,distserve,splitwise}, reuse attention state across requests~\cite{cacheblend}, and route queries to cheaper models~\cite{frugalgpt,routellm,vllmsr}.
These systems make each LLM call cheaper or pick the model, while a semantic cache removes the call and runs in front of them, as our vLLM testbed shows (\S\ref{sec:serving}).

\subsection{What Per-Entry Caches Leave Unspent}
\label{sec:measure}

\begin{figure}[t]
\centering
\includegraphics[width=0.325\linewidth]{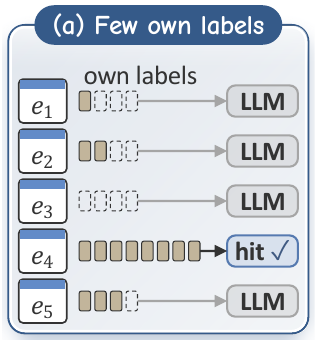}\hfill
\includegraphics[width=0.325\linewidth]{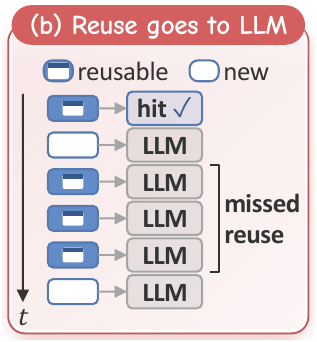}\hfill
\includegraphics[width=0.325\linewidth]{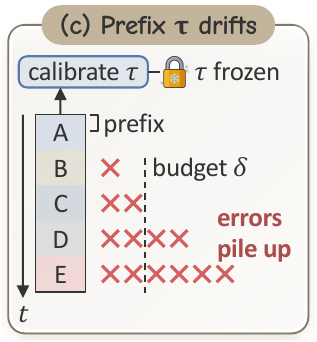}
\caption{Existing caches waste or break the error budget. (a) An entry serves hits only after four own labels. (b) Reusable queries still reach the LLM. (c) A prefix-calibrated threshold breaks the budget under drift.}
\Description{Three schematic cards. (a) Few own labels. Five cache entries hold one, two, zero, eight and three labels of their own in a row of four slots, and only the entry with eight labels serves a hit while the others call the LLM. (b) Reuse goes to the LLM. Six queries arrive over time, reusable ones marked with a cache card, and only the first reusable query hits while the next three reusable queries still call the LLM, marked as missed reuse. (c) Prefix threshold drifts. A threshold is calibrated on a prefix of class A and then frozen with a lock, and as classes B to E arrive the errors pile up past the budget line delta.}
\label{fig:motiv}
\end{figure}

\begin{figure}[t]
\centering
\includegraphics[width=\linewidth]{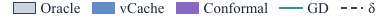}\\
\begin{subfigure}[t]{0.485\linewidth}\centering\includegraphics[width=\linewidth]{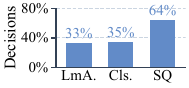}\caption{Few own labels}\end{subfigure}\hfill
\begin{subfigure}[t]{0.504\linewidth}\centering\includegraphics[width=\linewidth]{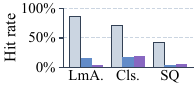}\caption{Unspent budget}\end{subfigure}\\[1pt]
\begin{subfigure}[t]{\linewidth}\centering\includegraphics[width=\linewidth]{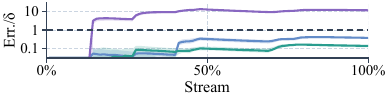}\caption{Drift on ClassificationSorted}\end{subfigure}
\caption{The three failures of Figure~\ref{fig:motiv} measured at $\delta=1\%$ on the test streams LmA., Cls. and SQ.}
\Description{Three measured panels. (a) Bars give the share of vCache decisions whose nearest entry has fewer than four labels of its own, 33, 35 and 64 percent on LmArena, Classification and SearchQueries. (b) Grouped bars give the hit rate of a per-entry hindsight oracle, vCache and Conformal on the three streams, with the oracle several times higher. (c) Lines give the cumulative error over delta along ClassificationSorted for vCache, GD and Conformal on a log scale, with Conformal rising far above the dashed line at one while vCache and GD stay below it.}
\label{fig:measured}
\end{figure}

\begin{figure*}[t]
\centering
\includegraphics[width=\textwidth]{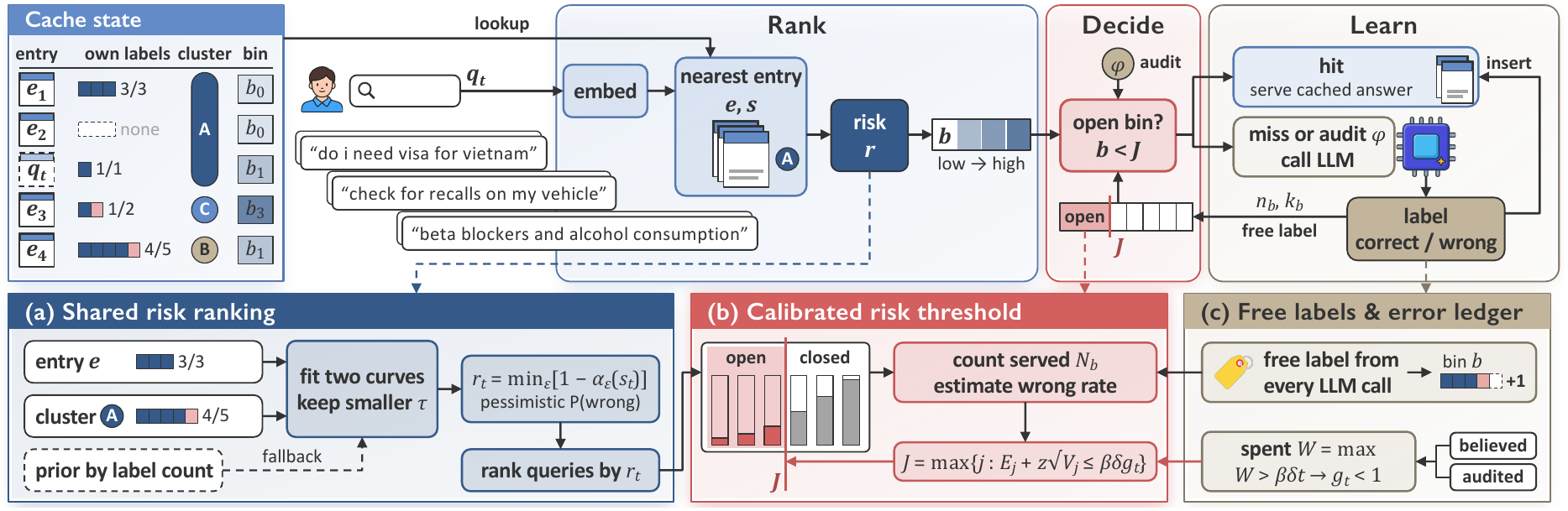}
\caption{Overview of \sys. Each query gets a shared risk score and is served from an open risk bin or sent to the LLM, whose answers become free labels.}
\Description{System diagram. Top left, a cache state table lists each entry with a short segmented bar and a count of its own labels, dark for correct and light red for wrong, its cluster and its risk bin, and a dashed row shows a query that an LLM call inserted into its cluster. Three bands follow. Rank embeds a query, finds its nearest entry with similarity s and its cluster, scores a pessimistic risk r and places the query in a risk bin ordered from low to high risk. Decide serves the cached answer when the bin is open, that is below the largest open bin J, and an audit coin sends some open-bin queries to the LLM. Learn calls the LLM on a miss or an audit, and the answer gives a correct or wrong label, a free label that updates the counts of the bin and is inserted into the cache. Below, panel a fits curves to the labels of an entry and of its cluster, falls back to a prior by the count of correct labels, keeps the smaller threshold, gives the pessimistic risk r and ranks queries by it. Panel b shows the bins, open or closed at J, each with a bar for its estimated wrong rate that grows with risk, counts the served decisions and estimates the wrong rate of each bin, and sets J as the most bins whose upper bound on the served error stays within the budget. Panel c turns every LLM call into a free label for its bin, drawn as one more cell on the count bar of bin b, and takes the spent error W as the larger of the believed and the audited errors, and when W runs ahead of the budget the factor g drops below one and tightens J.}
\label{fig:overview}
\end{figure*}

We replay the three shuffled test streams of \S\ref{sec:setup} and the sorted stream ClassificationSorted at $\delta=1\%$.
For vCache, its global dynamic threshold (GD) and Conformal, we log every cache decision with its nearest entry and the labels that this entry holds.

\paragraph{Many decisions meet an entry with too few labels.}
vCache serves no hit while the nearest entry has fewer than four labels (Figure~\ref{fig:motiv}a), and 33--64\% of its decisions fall there (Figure~\ref{fig:measured}a).
\emph{Evidence stays private to each entry}, which motivates one risk score shared by all paraphrases of a query (\S\ref{sec:rank}).

\paragraph{Much of the budget is left unspent.}
vCache stays within $\delta$ on the shuffled streams but still sends reusable queries to the LLM (Figure~\ref{fig:motiv}b). A hindsight oracle that picks one threshold per entry with the labels known reaches at least four times its hit rate at the same budget (Figure~\ref{fig:measured}b).
\emph{Charging $\delta$ to every query separately sends reusable queries to the LLM}, which motivates one threshold that pools the budget over a stream (\S\ref{sec:threshold}).

\paragraph{A threshold calibrated once does not stay safe.}
Conformal explores on the first 10\% of the stream and freezes a global threshold that holds $\delta$ on that prefix.
On ClassificationSorted, where the classes arrive in order (Figure~\ref{fig:motiv}c), its cumulative error ends at 11.38--12.53 times $\delta$ over five seeds, while vCache and GD stay below $\delta$ (Figure~\ref{fig:measured}c).
\emph{A drifting stream needs labels and an error check that keep coming}, which motivates the free labels and the error ledger of \S\ref{sec:ledger}.

\section{Surplus}
\label{sec:design}

\sys turns the \unspent into hits. It keeps the embedding, the index and the LLM of a standard semantic cache and changes how it decides and learns (Figure~\ref{fig:overview}).
Each decision computes a risk score from a shared model (\S\ref{sec:rank}) and serves the query if its risk bin is open under the calibrated threshold (\S\ref{sec:threshold}), and both address C1.
Every LLM call then labels a bin and updates an error ledger (\S\ref{sec:ledger}), which addresses C2.

\subsection{Preliminaries}
\label{sec:formulation}

\paragraph{Semantic cache.}
Queries $q_1,q_2,\dots$ arrive one at a time and are embedded by a fixed sentence encoder~\cite{sbert}.
For query $q_t$ the cache retrieves the most similar cached entry $e_t$ with cosine similarity $s_t$ and either serves its answer, a \emph{hit}, or calls the LLM, a \emph{miss}.
On a miss the LLM's answer also reveals whether the cached answer would have been correct, $c_t\in\{0,1\}$, and the cache may insert $q_t$ as a new entry.
Let $h_t\in\{0,1\}$ mark a hit.
Over a stream of $T$ queries the cache must keep its error rate within the budget:
\begin{equation}
\frac{1}{T}\sum_{t=1}^{T} h_t\,(1-c_t)\;\le\;\delta .
\label{eq:budget}
\end{equation}
As in vCache~\cite{vcache}, $\delta$ bounds the per-query error rate under the correctness signal that the cache receives on its misses.
The hit rate $\frac{1}{T}\sum_t h_t$ is what the cache maximizes.

\paragraph{Per-entry decisions.}
vCache fits for every entry a logistic curve $\Pr(c=1\mid s)=\sigma(\gamma(s-\theta))$ to the labels observed on that entry, plus two pseudo-labels at $s=0$ and $s=1$, and takes for each confidence level $1-\epsilon$ a pessimistic correctness bound $\alpha_\epsilon(s)$.
It then serves a hit when a uniform draw $u$ exceeds a threshold:
\begin{equation}
\tau(s)=\min_{\epsilon}\Big[1-\frac{\delta}{1-\alpha_\epsilon(s)}\Big],
\label{eq:tau}
\end{equation}
so that each query's error probability is at most $\delta$ when the curve family is right and the queries are i.i.d.
An entry needs several labels of its own before $\tau<1$, and every query is charged $\delta$ separately, whatever the risk of its neighbors.

\subsection{Shared Risk Ranking}
\label{sec:rank}
To address C1, \sys first ranks queries by how likely their reuse is wrong, with evidence shared among paraphrases.
When an LLM call confirms that a cached answer is correct for a query with similarity below an admission level $a$, \sys inserts the query as a new entry of the same \emph{cluster} of paraphrases, and a wrong answer starts a new cluster.
Every label $(s_t,c_t)$ is stored with the entry and with its cluster, so a new paraphrase inherits the evidence of its siblings in the cluster.

\paragraph{Risk score.}
\sys fits the logistic curve of (\ref{eq:tau}) to an entry's labels and to its cluster's labels once each has enough of them, refits a curve only after its label count grows by a fraction $\rho$, and keeps the smaller of the two $\tau$.
An entry whose cluster has no curve yet falls back to a prior pooled over all clusters with the same number of correct labels~\cite{efronmorris}, in three strata of 0, 1 to 3, and 4 or more, computed with a budget share of $0.65\delta$.
The risk score is the curve's pessimistic probability that the cached answer is wrong:
\begin{equation}
r_t=\frac{\delta_0}{1-\tau(s_t)}=\min_{\epsilon}\big[1-\alpha_\epsilon(s_t)\big],
\label{eq:risk}
\end{equation}
where $\delta_0$ is $\delta$ for an entry or cluster curve and $0.65\delta$ for the prior, whose $\tau$ takes $\delta_0$ in place of $\delta$ in (\ref{eq:tau}).
\sys uses $r_t$ only to rank queries, and the threshold calibrates its scale from the labels.

\subsection{Calibrated Risk Threshold}
\label{sec:threshold}
\sys then decides how far down the ranking to serve.
It places $r_t$ in one of 48 logarithmic bins, $b_t=\lfloor 12(\log_{10} r_t+4)\rfloor$ clipped to $[0,47]$, and serves the query if $b_t<J$, where $J$ is the number of open bins.
With probability $\phi$ such a query is \emph{audited} instead, so \sys calls the LLM and counts the query as a miss.

\paragraph{Served error per bin.}
\sys counts the decisions $N_b$ in each bin exactly, and the labels $n_b$ and wrong labels $k_b$ that each bin has received, all weighted by an exponential recency window of 3,000 decisions.
A pooled factor $\kappa=(\sum_b k_b+1)/(\sum_b n_b r_b+1)$, clipped to $[0.2,50]$, rescales the model risk of bin center $r_b$ into a prior, and a Beta posterior with strength $m$ gives the wrong rate of bin $b$:
\begin{equation}
\hat p_b=\frac{k_b+m\min(1,\kappa r_b)}{n_b+m}.
\label{eq:qb}
\end{equation}
Opening the first $j$ bins serves $(1-\phi)N_b$ queries of each, so after $T$ decisions it estimates the served error rate and its variance:
\begin{equation}
E_j=\frac{1-\phi}{T}\sum_{b<j}N_b\,\hat p_b,\qquad V_j=\frac{(1-\phi)^2}{T^2}\sum_{b<j}\frac{N_b^2\,\hat p_b(1-\hat p_b)}{n_b+m+1}.
\label{eq:ej}
\end{equation}

\paragraph{Threshold.}
Every 100 decisions \sys sets $J$:
\begin{equation}
J=\max\big\{\,j : E_j+z\sqrt{V_j}\le\beta\,\delta\,g_t\,\big\},
\label{eq:J}
\end{equation}
where $\beta<1$ keeps a margin, $z$ widens the estimate to an upper bound and $g_t\le 1$ comes from the ledger of \S\ref{sec:ledger}.
$J$ opens by at most three bins per update and closes at once, and the last bin opens in part when the bound leaves room.

\paragraph{Pooling the budget across queries is what spends it.}
When $\hat p_b$ grows with $b$, opening bins in order of risk serves the most queries for a given error mass, so low-risk paraphrases carry the budget that per-query rules leave idle.
The threshold needs no calibrated model, because $N_b$ is exact and only the wrong rate per bin is estimated, from labels the cache collects anyway.

\subsection{Free Labels and Error Ledger}
\label{sec:ledger}
To address C2, \sys keeps the estimates of \S\ref{sec:threshold} current with free labels and checks the errors it actually spends with an error ledger.
Every decision that reaches the LLM, an audit or a query in a closed bin, reveals $c_t$ for the nearest entry and adds a \emph{free label} to the bin of $r_t$ at no extra LLM cost.
Closed bins therefore keep receiving labels, so \sys can open them as soon as their wrong rate falls, and the recency window lets $\hat p_b$ follow the stream as it drifts.

\paragraph{Error ledger.}
\sys tracks the errors it has spent in two ways, the believed errors $W^{\mathrm{led}}_t=\sum_{i\le t,\,h_i=1} \hat p_{b_i}$ and the audit estimate $W^{\mathrm{aud}}_t=\frac{1-\phi}{\phi}A_t$, where $A_t$ counts the audited queries whose cached answer was wrong.
With $W_t$ the larger of $W^{\mathrm{led}}_t$ and $W^{\mathrm{aud}}_t$, the ledger sets the factor:
\begin{equation}
g_t=\mathrm{clip}\Big(1+\frac{\beta\delta t-W_t}{\beta\delta H},\;g_{\min},\;g_{\max}\Big),
\label{eq:ledger}
\end{equation}
where the horizon $H$ sets how fast $g_t$ reacts and $g_{\min}$ and $g_{\max}$ bound it, so whenever the spent errors run ahead of the accrued budget $\beta\delta t$, $g_t$ drops below one and (\ref{eq:J}) closes bins.
The audit estimate needs no model and no stationarity.
\begin{proposition}
If each query in an open bin is audited with probability $\phi$ independently of the past, then $W^{\mathrm{aud}}_t$ minus the number of wrong answers served up to $t$ is a martingale with mean zero, on any stream order.
\end{proposition}
\begin{proof}
The proof is an instance of inverse probability weighting~\cite{horvitz}.
Let $X_i\in\{0,1\}$ mark that the $i$-th query in an open bin has a wrong cached answer and $U_i\in\{0,1\}$ that it is audited.
Then $\Pr(U_i=1\mid\mathcal{F}_{i-1},X_i)=\phi$, where $\mathcal{F}_{i-1}$ is everything observed before the draw.
The served wrong answers up to $t$ are $S_t=\sum_{i\le t}(1-U_i)X_i$ and the audit estimate is $W^{\mathrm{aud}}_t=\frac{1-\phi}{\phi}\sum_{i\le t}U_iX_i$.
With $M_t=W^{\mathrm{aud}}_t-S_t$, each step satisfies:
\begin{equation*}
\mathbb{E}[M_t-M_{t-1}\mid\mathcal{F}_{t-1},X_t]=X_t\Big(\frac{1-\phi}{\phi}\,\phi-(1-\phi)\Big)=0,
\end{equation*}
so $M_t$ is a martingale with $M_0=0$ and $\mathbb{E}[M_t]=0$.
The variance of $M_t$ grows with the number of wrong answers rather than with the stream length $t$.
\end{proof}

\paragraph{The ledger reacts to realized errors, not to the model.}
Conformal and per-entry rules trust a model fitted earlier, so a drift that makes the model optimistic overspends silently.
The ledger compares two separate views of the spent errors with the budget accrued so far, and a stream that turns risky closes the threshold before the errors can pile up.

\subsection{Cost}
\label{sec:cost}
A decision adds one bin lookup to the nearest-neighbor search, the controller solves (\ref{eq:J}) over 48 bins every 100 decisions, and curves are refitted lazily.
Audits cost one LLM call for a share $\phi$ of the would-be hits and are counted as misses in every result we report.
Algorithm~\ref{alg:spend} gives the full loop, and Table~\ref{tab:hyper} in Appendix~\ref{app:impl} lists all settings.
The prior budget share, the strata and the admission level come from the shared risk model, and the threshold starts nearly closed, at the bins below a tenth of $\beta\delta$.

\begin{algorithm}[t]
\caption{\sys for one query $q_t$}
\label{alg:spend}
\small
\begin{algorithmic}[1]
\State $e_t,s_t\gets$ nearest cached entry and its similarity
\State $\tau\gets\min(\tau_{\mathrm{entry}},\tau_{\mathrm{cluster}})$ by (\ref{eq:tau}), or the stratified prior
\State $r_t\gets\delta_0/(1-\tau)$, $b_t\gets\lfloor 12(\log_{10}r_t+4)\rfloor$, $N_{b_t}\gets N_{b_t}+1$
\If{$b_t<J$ \textbf{and} a draw $u>\phi$}
  \State serve the answer of $e_t$, $W^{\mathrm{led}}\gets W^{\mathrm{led}}+\hat p_{b_t}$
\Else
  \State call the LLM, observe $c_t$, $n_{b_t}\gets n_{b_t}+1$, $k_{b_t}\gets k_{b_t}+1-c_t$
  \State \textbf{if} $b_t<J$ \textbf{and} $c_t=0$ \textbf{then} $A\gets A+1$
  \State update the entry and cluster labels, insert $q_t$ by the admission rule
\EndIf
\If{$t\bmod 100=0$}
  \State apply the recency window to the counts, update $\kappa$ and $\hat p_b$ by (\ref{eq:qb}), $g_t$ by (\ref{eq:ledger}) and $J$ by (\ref{eq:J})
\EndIf
\end{algorithmic}
\end{algorithm}

\section{Evaluation}
\label{sec:eval}

\begin{table*}[t]
\caption{Test hit rate (\%) per stream and budget.}
\label{tab:main}
\small
\setlength{\tabcolsep}{1.6pt}
\begin{tabular*}{\textwidth}{@{\extracolsep{\fill}}lrrrrrrrrrrrrrrrrr@{}}
\toprule
& \multicolumn{9}{c}{Shuffled streams} & \multicolumn{6}{c}{Drifting streams} & & \\
\cmidrule(lr){2-10}\cmidrule(lr){11-16}
& \multicolumn{3}{c}{LmArena} & \multicolumn{3}{c}{Classification} & \multicolumn{3}{c}{SearchQueries} & \multicolumn{3}{c}{ClassificationSorted} & \multicolumn{3}{c}{Combo} & & Worst \\
\cmidrule(lr){2-4}\cmidrule(lr){5-7}\cmidrule(lr){8-10}\cmidrule(lr){11-13}\cmidrule(lr){14-16}
Method \hfill $\delta$ (\%) & 0.5 & 1 & 2 & 0.5 & 1 & 2 & 0.5 & 1 & 2 & 0.5 & 1 & 2 & 0.5 & 1 & 2 & Held & err./$\delta$ \\
\midrule
GPTCache 0.97~\cite{gptcache} & \textcolor{black!40}{65.612} & \textcolor{black!40}{65.612} & \textcolor{black!40}{65.612} & \underline{15.947} & 15.947 & 15.947 & \textcolor{black!40}{23.197} & \textcolor{black!40}{23.197} & \textcolor{black!40}{23.197} & 16.260 & 16.260 & \underline{16.260} & 0.324 & 0.324 & 0.324 & 9/15 & 10.326 \\
GPTCache 0.98~\cite{gptcache} & \textcolor{black!40}{51.507} & \textcolor{black!40}{51.507} & \textcolor{black!40}{51.507} & 15.411 & 15.411 & 15.411 & \textcolor{black!40}{16.988} & \textcolor{black!40}{16.988} & \textcolor{black!40}{16.988} & 15.613 & 15.613 & 15.613 & 0.251 & 0.251 & 0.251 & 9/15 & 7.990 \\
GPTCache 0.99~\cite{gptcache} & \textcolor{black!40}{32.335} & \textcolor{black!40}{32.335} & \textcolor{black!40}{32.335} & 14.803 & 14.803 & 14.803 & \textcolor{black!40}{8.558} & \textcolor{black!40}{8.558} & 8.558 & 14.933 & 14.933 & 14.933 & 0.135 & 0.135 & 0.135 & 10/15 & 5.173 \\
Static~\cite{gptcache} & \textcolor{black!40}{13.534} & \textcolor{black!40}{13.534} & \textcolor{black!40}{32.335} & \textcolor{black!40}{27.823} & \textcolor{black!40}{33.565} & \textcolor{black!40}{40.944} & \textcolor{black!40}{8.558} & \textcolor{black!40}{16.988} & \textcolor{black!40}{23.197} & -- & -- & -- & -- & -- & -- & 0/9 & 2.706 \\
Krites~\cite{krites} & \textcolor{black!40}{13.534} & \textcolor{black!40}{13.534} & \textcolor{black!40}{37.819} & \textcolor{black!40}{23.381} & \textcolor{black!40}{35.897} & \textcolor{black!40}{35.897} & \textcolor{black!40}{13.145} & \textcolor{black!40}{18.651} & \textcolor{black!40}{25.615} & 14.200 & 14.200 & 14.973 & 0.058 & 0.058 & 0.135 & 6/15 & 2.858 \\
SISO~\cite{siso} & \textcolor{black!40}{3.548} & \textcolor{black!40}{5.563} & \textcolor{black!40}{7.790} & \textcolor{black!40}{27.823} & \textcolor{black!40}{33.565} & \textcolor{black!40}{40.944} & \textcolor{black!40}{4.182} & 4.182 & \textcolor{black!40}{9.681} & 10.449 & 10.449 & 12.538 & 0.164 & 0.164 & 0.269 & 7/15 & 2.706 \\
GD~\cite{vcache} & \textcolor{black!40}{6.590} & \textcolor{black!40}{13.961} & \textcolor{black!40}{42.330} & 1.031 & 2.048 & 4.060 & \textcolor{black!40}{1.153} & \textcolor{black!40}{2.328} & \textcolor{black!40}{4.592} & 3.112 & 6.116 & 11.124 & \textcolor{black!40}{0.960} & \textcolor{black!40}{1.952} & \textcolor{black!40}{3.940} & 6/15 & 2.180 \\
Conformal~\cite{crc} & 1.663 & 3.625 & 9.406 & 8.076 & \underline{19.208} & 26.432 & \underline{2.323} & \underline{5.370} & \underline{10.878} & \textcolor{black!40}{49.339} & \textcolor{black!40}{61.765} & \textcolor{black!40}{72.804} & 0.411 & 0.974 & 2.042 & 12/15 & 18.049 \\
vCache~\cite{vcache} & \underline{8.374} & \underline{16.350} & \underline{30.494} & 9.819 & 17.424 & \underline{33.701} & 1.893 & 3.481 & 6.556 & \underline{19.571} & \textbf{38.886} & \textcolor{black!40}{79.530} & \underline{3.964} & \underline{7.575} & \underline{15.533} & \underline{14/15} & \underline{1.430} \\
\midrule
\textbf{\sys} & \textbf{22.141} & \textbf{25.507} & \textbf{32.438} & \textbf{17.472} & \textbf{25.483} & \textbf{33.938} & \textbf{3.791} & \textbf{6.877} & \textbf{11.859} & \textbf{25.711} & \underline{36.370} & \textbf{47.708} & \textbf{5.221} & \textbf{9.493} & \textbf{17.244} & \textbf{15/15} & \textbf{0.840} \\
\bottomrule
\end{tabular*}
\par\smallskip
\begin{minipage}{\textwidth}
Gray numbers mark cells where a method breaks $\delta$ in some run.
\end{minipage}
\end{table*}

\subsection{Setup}
\label{sec:setup}

\paragraph{Streams.}
We use the three streams of the vCache benchmark~\cite{vcache}.
LmArena holds user prompts of Chatbot Arena~\cite{chatbotarena} and LLM paraphrases of them, Classification holds classification prompts, and SearchQueries holds Web search queries from ORCAS~\cite{orcas}.
Correctness labels come from the benchmark, equivalence classes for LmArena and SearchQueries and the answer string for Classification.
Each stream is split by a hash of the prompt into 20\% for validation, where methods with free settings pick them, and 80\% for test, which is 51,037, 36,000 and 120,000 queries.
We replay five random orders of each shuffled stream, and the drifting streams ClassificationSorted, with 45,000 queries sorted by class, and Combo, with 27,500 queries that mix several datasets~\cite{vcache}, in the benchmark's own order with five seeds.

\paragraph{Protocol.}
A method holds $\delta$ in a cell if its error rate is at most $\delta$ in every run of that cell, with $\delta\in\{0.5\%,1\%,2\%\}$.
Differences are paired per query, and 95\% confidence intervals (CIs) come from a nested bootstrap~\cite{efron79} that resamples runs and then queries.
A pooled gain averages the paired differences over the cells of each test order and then over the orders.
A difference is significant when its CI excludes zero.
In Table~\ref{tab:main}, a dash marks a setting not run.
Bold and underline mark the best and second best value among the methods that hold $\delta$ in each column.
The last two columns count the cells a method holds and give its worst error over $\delta$ in any run.

\paragraph{Baselines.}
GPTCache~\cite{gptcache} serves above a fixed similarity of 0.97, 0.98 or 0.99, and Static serves above the threshold with the most validation hits whose error stays within $\delta$ in every validation order, so it has no entry on the drifting streams, which are replayed whole without a validation split.
Krites~\cite{krites} serves above a high threshold and, below it, reuses entries whose answer an asynchronous judge has approved, with the benchmark labels as the judge. SISO~\cite{siso} serves from centroids of similar queries above a threshold.
GD is the global dynamic threshold of vCache~\cite{vcache}, and Conformal explores on the first 10\% of the stream and freezes the threshold of split conformal risk control~\cite{crc,gentleconf}.
vCache~\cite{vcache} is reimplemented from its official code, and MVR-cache~\cite{mvrcache} runs from its own code in \S\ref{sec:general}.
Appendix~\ref{app:setup} gives the answers, embeddings and index of each stream, the grid and pick of every baseline and the serving testbed.

\subsection{Main Results}
\label{sec:main}

\paragraph{\sys holds the budget in every test cell.}
It stays within $\delta$ in all runs of all cells with a worst error of 0.840 times $\delta$, while every other method in Table~\ref{tab:main} breaks it in some cell, shown in gray.
vCache breaks it in one cell, and Conformal breaks it by up to 18.049 times on ClassificationSorted.

\paragraph{\sys reuses more than every baseline that holds the budget on average.}
Pooled over the shuffled streams it adds 5.71 points over vCache and 4.10 points over the best baseline that holds $\delta$ in each cell (Table~\ref{tab:main} and Table~\ref{tab:cells}).
It is also significantly above 71 of the 73 baseline cells that hold $\delta$ (Table~\ref{tab:main}).
Table~\ref{tab:cells} in Appendix~\ref{app:cells} lists the best baseline that holds $\delta$ in each cell and the gain of \sys over it with its 95\% CI.
The largest gain is 31.45 points over GPTCache 0.97 on ClassificationSorted at $\delta=2\%$.

\paragraph{\sys beats every baseline that holds the budget on the shuffled streams and Combo.}
In every cell of these four streams, \sys has a higher hit rate than every baseline that holds $\delta$ there (Table~\ref{tab:main}).

\begin{figure}[t]
\centering
\includegraphics[width=\linewidth]{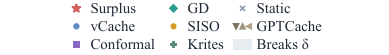}\\
\begin{subfigure}[t]{0.495\linewidth}\centering\includegraphics[width=\linewidth]{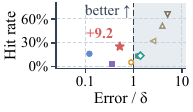}\caption{LmArena}\end{subfigure}\hfill
\begin{subfigure}[t]{0.495\linewidth}\centering\includegraphics[width=\linewidth]{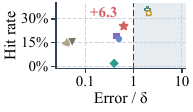}\caption{Classification}\end{subfigure}\\[1pt]
\begin{subfigure}[t]{0.495\linewidth}\centering\includegraphics[width=\linewidth]{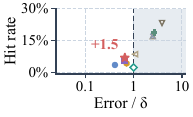}\caption{SearchQueries}\end{subfigure}\hfill
\begin{subfigure}[t]{0.495\linewidth}\centering\includegraphics[width=\linewidth]{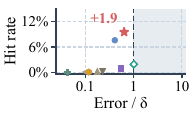}\caption{Combo}\end{subfigure}
\caption{Test hit rate against the share of $\delta$ spent at $\delta=1\%$, one marker per method of Table~\ref{tab:main}.}
\Description{Four scatter panels of hit rate against mean error over delta on a log axis, with a dashed line at one. In each panel the Surplus star is the highest marker left of the line, the baselines that hold delta lie further left, and the fixed thresholds with more hits lie right of the line with hollow markers.}
\label{fig:budget}
\end{figure}

\begin{figure*}[t]
\centering
\includegraphics[width=\textwidth]{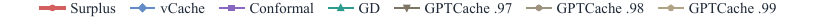}\\
\begin{subfigure}[t]{0.264\linewidth}\centering\includegraphics[width=\linewidth]{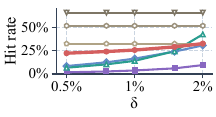}\caption{LmArena}\end{subfigure}\hfill
\begin{subfigure}[t]{0.241\linewidth}\centering\includegraphics[width=\linewidth]{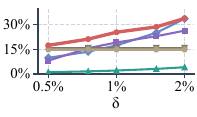}\caption{Classification}\end{subfigure}\hfill
\begin{subfigure}[t]{0.241\linewidth}\centering\includegraphics[width=\linewidth]{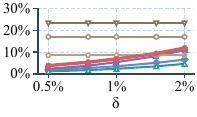}\caption{SearchQueries}\end{subfigure}\hfill
\begin{subfigure}[t]{0.241\linewidth}\centering\includegraphics[width=\linewidth]{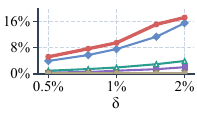}\caption{Combo}\end{subfigure}
\caption{Test hit rate of \sys, vCache, Conformal, GD and the three GPTCache thresholds across five error budgets, where hollow markers break $\delta$ in some run.}
\Description{Four panels of hit rate against the error budget for Surplus, vCache, Conformal, GD and the three GPTCache thresholds, one each for LmArena, Classification, SearchQueries and Combo.}
\label{fig:curves_all}
\end{figure*}

\paragraph{\sys spends the budget others leave.}
At $\delta=1\%$ on these four streams, most baselines that hold $\delta$ spend less than half of it, and every method that reuses more than \sys breaks it (Figure~\ref{fig:budget}).
Each marker gives a method's mean error over $\delta$ across runs. Methods right of the dashed line break $\delta$ on average, and a hollow marker breaks it in some run.

\paragraph{\sys leads at every budget on four streams.}
On a finer grid of five budgets, \sys holds $\delta$ at all 20 points of these four streams and lies above the best baseline that holds $\delta$ at each point (Figure~\ref{fig:curves_all}).
Pooled over the shuffled streams its lead is 4.32 points (Table~\ref{tab:cells}), and it is largest at tight budgets on LmArena.

\subsection{Where the Gain Comes From}
\label{sec:ablation}

\begin{table}[t]
\caption{Hit-rate change in points when parts of \sys are removed, pooled over shuffled or drifting streams, and the worst error over $\delta$, where values above 1 break $\delta$ and dashes mark streams not run.}
\label{tab:ctrl}
\small
\setlength{\tabcolsep}{3.2pt}
\begin{tabular*}{\linewidth}{@{\extracolsep{\fill}}lrrrr}
\toprule
& \multicolumn{2}{c}{Hit change} & \multicolumn{2}{c}{Worst err./$\delta$} \\
\cmidrule(lr){2-3}\cmidrule(lr){4-5}
Variant & Shuffled & Drifting & Shuffled & Drifting \\
\midrule
w/o cluster sharing & $-$2.82 & -- & 0.710 & -- \\
w/o stratified prior & $-$0.645 & -- & 0.722 & -- \\
w/o admission & $-$1.34 & +5.72 & 0.693 & 0.680 \\
w/o free labels & $-$0.900 & +0.137 & 0.733 & 0.816 \\
w/o ledger & +0.412 & +1.45 & 0.765 & 0.924 \\
w/o free labels \& ledger & -- & +1.50 & -- & 1.227 \\
\midrule
w/o threshold, full refit & $-$2.78 & +4.81 & 1.015 & 2.294 \\
w/o threshold & $-$2.79 & +4.93 & 1.022 & 2.833 \\
\midrule
\textbf{\sys} & 0.000 & 0.000 & 0.736 & 0.840 \\
\bottomrule
\end{tabular*}
\end{table}

\begin{figure}[t]
\centering
\includegraphics[width=\linewidth]{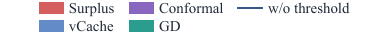}\\
\begin{subfigure}[t]{0.490\linewidth}\centering\includegraphics[width=\linewidth]{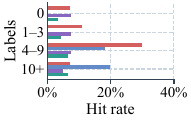}\caption{Hits by entry labels}\end{subfigure}\hfill
\begin{subfigure}[t]{0.499\linewidth}\centering\includegraphics[width=\linewidth]{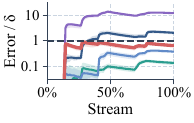}\caption{ClassificationSorted}\end{subfigure}
\caption{\sys serves sparsely labeled entries and keeps the error under $\delta$ on a sorted stream where Conformal and \sys without the threshold drift above it, at $\delta=1\%$.}
\Description{Left, bars of hit rate by the number of labels of the nearest entry for Surplus, vCache, Conformal and GD, pooled over the shuffled streams, where vCache has no hits below four labels and Surplus has hits in every bucket. Right, cumulative error over the budget along ClassificationSorted on a log axis for GD, vCache, Conformal, Surplus without the threshold and Surplus, with bands over five seeds, where Conformal and the control rise above one and Surplus stays below it.}
\label{fig:behavior}
\end{figure}

\begin{figure}[t]
\centering
\includegraphics[width=\linewidth]{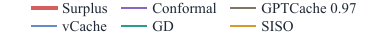}\\
\begin{subfigure}[t]{0.336\linewidth}\centering\includegraphics[width=\linewidth]{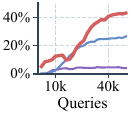}\caption{LmArena}\end{subfigure}\hfill
\begin{subfigure}[t]{0.336\linewidth}\centering\includegraphics[width=\linewidth]{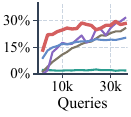}\caption{Classification}\end{subfigure}\hfill
\begin{subfigure}[t]{0.317\linewidth}\centering\includegraphics[width=\linewidth]{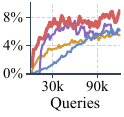}\caption{SearchQueries}\end{subfigure}
\caption{Hit rate in consecutive windows of 2,000 queries at $\delta=1\%$, averaged over the test orders, for the methods that hold $\delta$ on each stream.}
\Description{Three line charts of hit rate per window along the stream for Surplus and the baselines that hold the budget on each stream, vCache and Conformal on all three and also GD and GPTCache 0.97 on Classification. The Surplus line rises earlier and lies above the vCache line in most windows.}
\label{fig:coldstart}
\end{figure}

\paragraph{Shared evidence turns sparse entries into hits.}
Pooled over the shuffled streams, \sys serves hits from entries with no label of their own and with one to three labels, where vCache serves none (Figure~\ref{fig:behavior}a).
Without cluster sharing the hit rate drops by 2.82 points, and without the stratified prior or free labels it also drops (Table~\ref{tab:ctrl}).
Every entry is sparse early in a stream, and \sys starts reusing earlier than vCache.
In the first 2,000 queries of each shuffled stream it already serves more hits than vCache, and it stays ahead in most later windows (Figure~\ref{fig:coldstart}).

\paragraph{The calibrated threshold is what keeps the budget.}
Two controls keep the shared risk model of \sys but use vCache's rule (\ref{eq:tau}) in place of the threshold, with the lazy refit of \sys or with a refit at every label.
Both spend $\delta$ per query and break $\delta$ on the drifting streams, by up to 2.833 times (Table~\ref{tab:ctrl}).
On ClassificationSorted the lazy-refit control climbs above $\delta$ as the classes change, while \sys stays below it (Figure~\ref{fig:behavior}b).

\paragraph{Free labels and the ledger back each other up.}
Removing either one keeps $\delta$ on both drifting streams, while removing both breaks $\delta$ on ClassificationSorted with a worst error of 1.227 times $\delta$ (Table~\ref{tab:ctrl}).

\subsection{Web Search Queries}
\label{sec:web}

\begin{figure}[t]
\centering
\includegraphics[width=\linewidth]{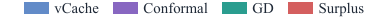}\\
\begin{subfigure}[t]{0.495\linewidth}\centering\includegraphics[width=\linewidth]{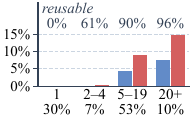}\caption{By paraphrase class size}\end{subfigure}\hfill
\begin{subfigure}[t]{0.495\linewidth}\centering\includegraphics[width=\linewidth]{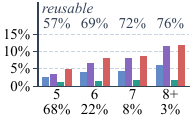}\caption{By query length}\end{subfigure}
\caption{Correct hits per bucket of SearchQueries at $\delta=1\%$, with each bucket's share of the stream under its label and its reusable share above its bars.}
\Description{Two vertical bar charts side by side over SearchQueries buckets. By class size, singleton queries are 30 percent of the stream and never reusable, while classes of 20 or more queries are 96 percent reusable and Surplus serves about twice vCache's correct hits there. By query length, with Conformal and GD added, reusable shares rise from 57 percent for five-word queries to 76 percent for eight or more words, and Surplus leads vCache in every bucket.}
\label{fig:web}
\end{figure}

\begin{figure}[t]
\centering
\includegraphics[width=\linewidth]{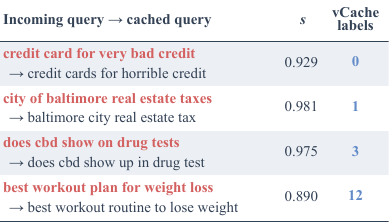}
\caption{Search queries that \sys serves from the cache and vCache sends to the LLM at $\delta=1\%$, all judged correct by a human reviewer.}
\Description{Four pairs of an incoming search query and the cached query it reuses, for example credit card for very bad credit and credit cards for horrible credit, with their cosine similarity and the number of labels the matched entry had under vCache, between 0 and 12.}
\label{fig:examples}
\end{figure}

We split the SearchQueries test stream, which comes from ORCAS search logs, by the size of each query's paraphrase class and by query length, and count the correct hits that each cache serves per bucket at $\delta=1\%$ (Figure~\ref{fig:web}).

\paragraph{Reuse concentrates in popular paraphrase classes and longer queries.}
Queries whose class has 20 or more members in the stream are 96\% reusable, and there \sys serves nearly twice vCache's correct hits (Figure~\ref{fig:web}a).
Singleton queries make up 30\% of the stream and can never be reused, so a cache for search traffic benefits most from the head of the class distribution.
Queries of eight or more words are 76\% reusable against 57\% for five-word queries, and \sys's lead over vCache holds in every length bucket (Figure~\ref{fig:web}b).

\paragraph{\sys reuses paraphrases that vCache sends to the LLM\@.}
The queries in Figure~\ref{fig:examples} share an intent with a cached query in other words, and vCache sends them to the LLM because the matched entry has few labels or its curve is still uncertain at that similarity.

\subsection{Feedback Signals and Human Review}
\label{sec:judge}

We change the signal that labels the misses or grades the hits, with flipped labels, LLM judges~\cite{llmjudge} and human reviewers.
When 5\% or 10\% of the labels that each method receives are flipped on the validation split, \sys holds $\delta$ in all nine cells and keeps a higher pooled hit rate than vCache, 7.37\% against 4.91\% at 5\% and 5.13\% against 4.12\% at 10\% (Appendix~\ref{app:sens}).

\begin{table}[t]
\caption{Hit rate (\%) and error over $\delta$ on one LmArena test order when Qwen3.8-27B labels the misses and judges the hits, with the higher hit rate per budget in bold.}
\label{tab:judge27}
\small
\setlength{\tabcolsep}{2pt}
\begin{tabular*}{\linewidth}{@{\extracolsep{\fill}}lrrrrrr}
\toprule
& \multicolumn{2}{c}{$\delta=0.5\%$} & \multicolumn{2}{c}{$\delta=1\%$} & \multicolumn{2}{c}{$\delta=2\%$} \\
\cmidrule(lr){2-3}\cmidrule(lr){4-5}\cmidrule(lr){6-7}
Method & Hit & Err./$\delta$ & Hit & Err./$\delta$ & Hit & Err./$\delta$ \\
\midrule
vCache~\cite{vcache} & 1.64 & 0.513 & 3.00 & 0.460 & 6.06 & 0.409 \\
\textbf{\sys} & \textbf{2.53} & 0.482 & \textbf{4.82} & 0.652 & \textbf{7.87} & 0.641 \\
\bottomrule
\end{tabular*}
\end{table}

\paragraph{\sys leads under an LLM judge.}
When Qwen3.8-27B~\cite{qwen38} labels the misses and judges the hits on one LmArena test order, \sys and vCache hold $\delta$ at every budget (Table~\ref{tab:judge27}).
\sys serves significantly more hits than vCache at each budget, by 1.82 points at $\delta=1\%$.
This judge accepts fewer reuses than Qwen3-8B~\cite{qwen3}, so both caches serve 44--62\% of the hits they serve with Qwen3-8B on the same order.
With Qwen3-8B as the judge, both caches also hold $\delta$ in every cell of LmArena, Classification, SearchQueries and Quora question pairs (QQP)~\cite{glue} (Table~\ref{tab:r2b} in Appendix~\ref{app:human}).
On SearchQueries at $\delta=1\%$ under this judge, \sys serves 12.25\% of the queries against 5.36\% for vCache.

\paragraph{Under human review \sys serves more correct answers.}
In a blind review of hits from both methods, \sys serves 6.72 more correct answers per 100 LmArena queries than vCache.
Two annotators reach a Cohen's kappa of 0.957 on the search queries they both labeled~\cite{cohen}, and Appendix~\ref{app:impl} describes the protocol.
The budget is defined under the feedback signal (\S\ref{sec:formulation}), so it is only as strict as that signal, and on search queries the benchmark labels are looser than the judgments of human reviewers.

\subsection{Generality}
\label{sec:general}

\begin{figure}[t]
\centering
\includegraphics[width=\linewidth]{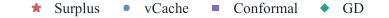}\\
\includegraphics[width=\linewidth]{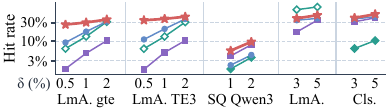}\\[1pt]
\includegraphics[width=0.495\linewidth]{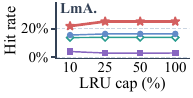}\hfill
\includegraphics[width=0.495\linewidth]{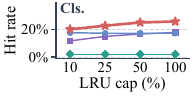}\\[1pt]
\includegraphics[width=0.495\linewidth]{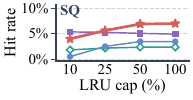}\hfill
\includegraphics[width=0.495\linewidth]{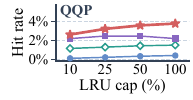}
\caption{Hit rate with other embeddings and budgets (top) and under an LRU cap at $\delta=1\%$ (bottom), where LmA., Cls. and SQ denote LmArena, Classification and SearchQueries.}
\Description{Top, hit rate on a log axis in five groups, LmArena with the gte and TE3-small embeddings at budgets of 0.5, 1 and 2 percent, SearchQueries with Qwen3-Embedding at 1 and 2 percent, and LmArena and Classification at 3 and 5 percent, where the Surplus star is the highest marker that holds delta in every group and GD markers are hollow on LmArena. Bottom, four panels of hit rate against an LRU cap of 10, 25, 50 and 100 percent of the stream on LmArena, Classification, SearchQueries and Quora QQP, where Surplus lies above vCache at every cap.}
\label{fig:general}
\end{figure}

\paragraph{The gain carries over to other embeddings and budgets.}
With gte~\cite{gte} and TE3-small~\cite{te3} on LmArena, Qwen3-Embedding~\cite{qwen3emb} on SearchQueries and the main embeddings at $\delta$ of 3\% and 5\%, \sys holds $\delta$ in all 12 cells and leads vCache by 9.96 points pooled (Table~\ref{tab:vgain}).
In each cell it is also significantly above every baseline that holds $\delta$ there, vCache, Conformal or GD (Figure~\ref{fig:general}, top).
On Quora question pairs~\cite{glue}, a fourth public stream, \sys holds $\delta$ and is significantly above vCache in all three cells.

\paragraph{The lead survives a capacity limit.}
With an LRU cache capped at 10\% to 100\% of the stream length, \sys holds $\delta$ in every run on all four streams.
It keeps a significant lead over vCache at every cap (Figure~\ref{fig:general}, bottom).
From a cap of a quarter of the stream, it has the highest hit rate of the methods that hold $\delta$ on every stream.
On SearchQueries this needs at least that cap, and eviction rules aware of the risk ranking are left for future work.
LmArena never fills a quarter of the stream, so only its smallest cap evicts entries.

\begin{figure}[t]
\centering
\includegraphics[width=\linewidth]{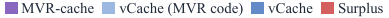}\\
\begin{subfigure}[t]{0.510\linewidth}\centering\includegraphics[width=\linewidth]{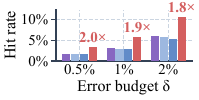}\caption{Hit rate}\end{subfigure}\hfill
\begin{subfigure}[t]{0.479\linewidth}\centering\includegraphics[width=\linewidth]{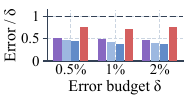}\caption{Worst error over $\delta$}\end{subfigure}
\caption{\sys against MVR-cache on the full SearchQueries stream with the bge-base embedding.}
\Description{Left, grouped bars of hit rate at budgets of 0.5, 1 and 2 percent for MVR-cache, vCache from the MVR-cache code, vCache and Surplus, where Surplus reaches 2.0, 1.9 and 1.8 times the hit rate of MVR-cache. Right, the worst error over delta of the same four, all below a dashed line at one.}
\label{fig:mvr}
\end{figure}

\paragraph{\sys gains more over vCache than a learned similarity does.}
MVR-cache~\cite{mvrcache} changes the similarity function and keeps vCache's decision, while \sys leaves the similarity untouched and changes the decision.
On three orders of the full SearchQueries test stream with the same bge-base embedding~\cite{bge}, \sys serves 1.8 to 2.0 times the hits of MVR-cache at every budget.
All four caches hold $\delta$ there (Figure~\ref{fig:mvr}).
On SearchQueries and Classification, its gain over the vCache of its own code base exceeds MVR-cache's gain over its vCache in every cell (Table~\ref{tab:mvr} in Appendix~\ref{app:cells}).
On SearchQueries at $\delta=1\%$, \sys serves 5.95\% against 2.87\% for its vCache, while MVR-cache serves 3.17\% against 3.01\% for its own.
The two code bases label Classification by different keys, so each method is compared with its own vCache.

\subsection{Cost and Serving}
\label{sec:serving}

\begin{figure}[t]
\centering
\includegraphics[width=\linewidth]{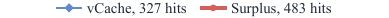}\\
\begin{subfigure}[t]{0.490\linewidth}\centering\includegraphics[width=\linewidth]{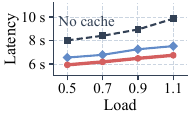}\caption{Poisson arrivals}\end{subfigure}\hfill
\begin{subfigure}[t]{0.490\linewidth}\centering\includegraphics[width=\linewidth]{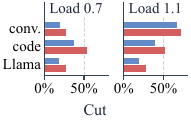}\caption{Traces and Llama}\end{subfigure}
\caption{Mean end-to-end latency of LmArena queries served by vLLM at $\delta=1\%$ and its cut from serving without a cache, where the load is a multiple of the closed-loop throughput of the backend.}
\Description{Left, a line chart of mean latency in seconds against the load under Poisson arrivals, where Surplus lies below vCache and both lie below serving without a cache. Right, horizontal bars of the latency cut from serving without a cache on the Azure conversation and code traces and with a Llama backend at two loads, where Surplus cuts more than vCache in every row. The legend gives 327 cache hits for vCache and 483 for Surplus.}
\label{fig:e2e}
\end{figure}

\paragraph{\sys has a lower decision latency than vCache on average.}
Its mean and p99 decision times are below vCache's on every stream (Table~\ref{tab:overhead} in Appendix~\ref{app:cells}).
On one CPU core its mean decision time is about a third of vCache's on LmArena and under a fifth on Classification, and its p99 on SearchQueries is under half of vCache's.
As the SearchQueries prefix grows to 120,000 queries its wall-clock time stays within 1.09 times vCache's (Appendix~\ref{app:sens}).

\paragraph{More hits mean lower latency in serving.}
vLLM~\cite{vllm} serves Qwen3-8B on one RTX 4090, and 2,000 LmArena queries arrive as a Poisson process at four loads up to 1.1 times its closed-loop throughput (Appendix~\ref{app:setup}).
\sys lowers the mean latency more than vCache at every load, by 31.59\% against 23.67\% at the highest load (Figure~\ref{fig:e2e}a).
It answers 483 of the 2,000 queries without the LLM where vCache answers 327 of them.

\paragraph{The latency gain holds under bursty arrivals and other backends.}
With arrivals replayed from the conversation and code traces of Azure LLM inference~\cite{splitwise} and with a Llama-3.1-8B backend~\cite{llama3}, \sys cuts the mean latency more than vCache in every setting (Figure~\ref{fig:e2e}b).
On the burstier code trace at 0.7 times the throughput, it cuts the mean latency by 53.09\% against 37.90\% for vCache.
With Qwen3.8-27B~\cite{qwen38} served on one A800 GPU at 0.7 times its throughput, \sys cuts the mean latency by 29.67\% and vCache by 20.94\% at the same load.

\section{Conclusion}
\label{sec:conclusion}

Semantic caches that bound the error per entry at the target budget, such as vCache, leave much of the error budget of LLM Web services unspent, and \sys spends it where reuse is most likely correct.
\sys ranks queries by a risk score shared across paraphrase clusters, opens risk bins under a threshold calibrated from audited labels, and keeps it safe under drift with free labels and a ledger of realized errors.
On search, chat and classification streams it holds the budget in all 15 test cells, which no baseline does, and serves 40\% more hits than vCache on the shuffled streams.
In a vLLM testbed it cuts the mean latency by 31.59\% at the highest load, against 23.67\% for vCache.
We expect the \unspent to matter wherever an LLM service promises a quality level, and future work will extend the ledger to caches shared by tenants with their own budgets.

\bibliographystyle{ACM-Reference-Format}
\bibliography{refs_gen,refs_manual}

\appendix
\section*{Appendix}
The appendix gives the streams, baselines and testbed (\ref{app:setup}), run and review protocols (\ref{app:impl}), per-cell results (\ref{app:cells}), noise and scaling (\ref{app:sens}), the judge feedback (\ref{app:human}) and the use of generative models (\ref{app:genai}).

\section{Streams, Baselines and Testbed}
\label{app:setup}

\paragraph{Streams.}
LmArena comes with GPT-4o-mini answers~\cite{gpt4o} and e5-large-v2 embeddings~\cite{e5}. Classification comes with Llama~3 answers~\cite{llama3}, and Classification and SearchQueries both come with gte embeddings~\cite{gte}.
The cache retrieves with HNSW~\cite{hnsw} and starts every run empty, the baselines insert every miss, and \sys inserts by the admission rule of \S\ref{sec:rank}.

\paragraph{Baseline settings.}
Static picks its threshold from 0.60 to 0.999 on the validation split, and when no threshold keeps every validation order within $\delta$ we report the largest one.
Krites and SISO take the setting with the most validation hits whose worst validation order stays within $\delta$.
On the drifting streams they take the setting that does so over all nine validation runs of the shuffled streams.
Where no setting stays within $\delta$, as for Krites on LmArena and the drifting streams at the two tighter budgets, we report the setting with the lowest worst error.
Krites picks its high and low thresholds from 0.80 to 0.999 and serves approved entries at either one, both in the grid.
SISO merges queries into centroids at the similarity of 0.86 from its paper.
Its latency-driven threshold stays fixed in a replay without load and is picked from 0.60 to 0.98.
Conformal labels the first 10\% of the stream by calling the LLM on every query and then serves above the threshold that conformal risk control picks at level $\delta$.
vCache runs with its error bound set to the target $\delta$ of each cell, and our vCache agrees with its official implementation on every query of a check stream.
MVR-cache trains its segmentation model on 3,000 validation prompts per stream with its own code.
It runs on one order of Classification and on three orders of the full SearchQueries test stream, and \sys and vCache run on the same data with the same bge-base embedding in our replay.
A hindsight oracle picks per-entry thresholds with all labels known (Figure~\ref{fig:measured}b).

\paragraph{Serving testbed.}
vLLM 0.11 serves Qwen3-8B in bf16 without thinking on one RTX 4090.
The client sends queries 20,000 to 21,999 of one LmArena test order after the cache has replayed the first 20,000.
Arrivals are Poisson at 0.5, 0.7, 0.9 and 1.1 times the backend's closed-loop throughput of 3.315 queries per second.
A miss generates as many tokens as the logged GPT-4o-mini answer, up to 1,024.
Hits and served entries follow the main-table traces, the live decisions are only timed, and the feedback labels come from the benchmark as in the main table.
For Qwen3.8-27B, vLLM 0.17 serves the model in bf16 on one A800 with a closed-loop throughput of 1.598 queries per second.
The same 2,000 queries arrive as a Poisson process, and the run without a cache uses a second A800 with the same server settings.
The trace runs replay the inter-arrival gaps of the Azure conversation trace, whose coefficient of variation is 1.14, scaled to 0.7 and 1.1 times the same throughput.
The Llama run serves Llama-3.1-8B-Instruct at 0.7 times its own closed-loop throughput of 4.264 queries per second.
The runs on the Azure code trace, whose gaps have a coefficient of variation of 13.33, and the Llama run at 1.1 times use a second server with one RTX 4090D.
These runs scale their load to the closed-loop throughputs of that server, 3.346 queries per second for Qwen3-8B and 4.348 for Llama-3.1-8B.
Appendix~\ref{app:impl} gives the run counts, the timing protocol and the review protocol.

\section{Run and Review Protocols}
\label{app:impl}

\paragraph{Runs and timing.}
The ablations of Table~\ref{tab:ctrl} run three test orders of LmArena and Classification, one of SearchQueries and two seeds of each drifting stream, and its rows without the threshold come from the full main-table runs.
The capacity runs of Table~\ref{tab:vgain} and the runs of Figure~\ref{fig:general} use two test orders of each stream, except one of SearchQueries with Qwen3-Embedding.
Timing runs give each method one CPU core of a shared server and run all methods of a cell at the same time, twice each.
The methods of a cell thus share one load, while the load differs across cells and between Table~\ref{tab:overhead} and the prefix runs, so we compare methods only within a cell.

\begin{table}[t]
\caption{Settings of \sys, fixed on the validation split and used unchanged for every stream, budget and test run.}
\label{tab:hyper}
\small
\begin{tabular*}{\linewidth}{@{\extracolsep{\fill}}ll}
\toprule
Setting & Value \\
\midrule
Risk bins & 48, 12 per decade from $10^{-4}$ \\
Control interval & 100 decisions \\
Audit rate $\phi$ & 0.2 \\
Target share $\beta$ & 0.75 \\
Bound width $z$ & 1.5 \\
Prior strength $m$ & 4 \\
Opening step & 3 bins per control \\
Recency window & 3,000 decisions \\
Ledger horizon $H$ & 1,500 decisions \\
Ledger clip $[g_{\min},g_{\max}]$ & $[0.25, 1]$ \\
Calibration clip of $\kappa$ & $[0.2, 50]$ \\
Admission $a$ & 0.9 \\
Lazy refit $\rho$ & 0.1 \\
Prior budget share & 0.65 \\
Strata of correct labels & 0, 1 to 3, 4 or more \\
\bottomrule
\end{tabular*}
\end{table}

\paragraph{Human reviews.}
Two members of our research group reviewed sampled hits of \sys at $\delta=1\%$.
The review page shows a Chinese machine translation by DeepSeek-V4-Flash~\cite{deepseekv4} at temperature 0 as a reference only, with the English original as the ground truth, and highlights the words that differ between the new and the cached query.
We checked every translation and corrected 6 by hand, and the two annotators reach a Cohen's kappa of 0.957 on the 49 SearchQueries items they both labeled and agree on all 15 Classification items they both labeled.
One annotator labeled the SearchQueries and Classification items, the other the LmArena items, all SearchQueries items and 15 Classification items, and the items labeled by both give the agreement.
A second review compares the two methods on 151 hits sampled evenly from queries that only \sys serves, that both serve and that only vCache serves, and weights each class by its size.
Both reviews draw their SearchQueries hits from two of its five test orders.

\section{Per-Cell Results, Generality and Cost}
\label{app:cells}

\begin{table}[tb]
\caption{Hit rate (\%) of vCache and \sys with other embeddings and budgets, on Quora QQP and under an LRU cap, with the gain in points and its nested bootstrap 95\% CI.}
\label{tab:vgain}
\small
\setlength{\tabcolsep}{1.4pt}
\begin{tabular*}{\linewidth}{@{\extracolsep{\fill}}lrrrrr@{}}
\toprule
Setting & $\delta$ (\%) & vCache & \textbf{\sys} & Gain & CI \\
\midrule
LmArena, gte & 0.5 & 9.17 & 27.09 & +17.92 & [17.46, 18.36] \\
 & 1 & 17.71 & 30.78 & +13.07 & [12.56, 13.57] \\
 & 2 & 32.55 & 36.80 & +4.25 & [3.85, 4.63] \\
LmArena, TE3-small & 0.5 & 11.04 & 35.21 & +24.16 & [23.85, 24.49] \\
 & 1 & 20.86 & 38.31 & +17.44 & [17.13, 17.76] \\
 & 2 & 37.94 & 43.66 & +5.72 & [5.22, 6.27] \\
SearchQueries, Qwen3-Emb. & 1 & 2.32 & 5.61 & +3.29 & [3.16, 3.44] \\
 & 2 & 4.32 & 9.54 & +5.22 & [5.05, 5.36] \\
LmArena, main & 3 & 35.33 & 39.49 & +4.15 & [3.24, 5.02] \\
 & 5 & 39.38 & 47.83 & +8.46 & [8.08, 8.84] \\
Classification, main & 3 & 38.11 & 40.73 & +2.62 & [0.31, 4.90] \\
 & 5 & 43.49 & 50.76 & +7.26 & [6.27, 8.31] \\
Pooled over these rows & & & & +9.96 & [9.65, 10.28] \\
\midrule
Quora QQP, gte & 0.5 & 0.23 & 2.18 & +1.95 & [1.79, 2.08] \\
 & 1 & 0.41 & 3.91 & +3.50 & [3.36, 3.62] \\
 & 2 & 0.80 & 6.26 & +5.46 & [5.26, 5.66] \\
\midrule
LmArena, LRU 10\% & 1 & 15.63 & 22.00 & +6.38 & [6.09, 6.65] \\
LmArena, LRU 25, 50\% & 1 & 16.45 & 25.25 & +8.80 & [8.52, 9.07] \\
Classification, LRU 10\% & 1 & 17.56 & 20.01 & +2.45 & [2.10, 2.81] \\
Classification, LRU 25\% & 1 & 17.17 & 22.60 & +5.44 & [5.07, 5.80] \\
Classification, LRU 50\% & 1 & 17.17 & 24.94 & +7.78 & [7.42, 8.17] \\
\bottomrule
\end{tabular*}
\end{table}

\begin{table}[tb]
\caption{Hit rate (\%) of MVR-cache and \sys with the same bge-base embedding, each beside the vCache of its own code, with the top two in bold and underline.}
\label{tab:mvr}
\small
\setlength{\tabcolsep}{3pt}
\begin{tabular*}{\linewidth}{@{\extracolsep{\fill}}lrrrrrr}
\toprule
& \multicolumn{3}{c}{Classification} & \multicolumn{3}{c}{SearchQueries} \\
\cmidrule(lr){2-4}\cmidrule(lr){5-7}
Method \hfill $\delta$ (\%) & 0.5 & 1 & 2 & 0.5 & 1 & 2 \\
\midrule
MVR-cache & \underline{12.27} & \underline{16.91} & \textbf{33.09} & \underline{1.75} & \underline{3.17} & \underline{6.06} \\
vCache, MVR code & 9.14 & 16.46 & \underline{32.49} & 1.67 & 3.01 & 5.75 \\
vCache, our code & 8.43 & 15.62 & 30.48 & 1.59 & 2.87 & 5.46 \\
\midrule
\textbf{\sys} & \textbf{13.26} & \textbf{19.43} & 31.13 & \textbf{3.49} & \textbf{5.95} & \textbf{10.68} \\
\bottomrule
\end{tabular*}
\end{table}

\begin{table}[tb]
\caption{Decision time in milliseconds, wall-clock time of a whole test run and peak memory at $\delta=1\%$, on one CPU core.}
\label{tab:overhead}
\small
\setlength{\tabcolsep}{3pt}
\begin{tabular*}{\linewidth}{@{\extracolsep{\fill}}llrrrrr}
\toprule
& & \multicolumn{3}{c}{Decision (ms)} & & \\
\cmidrule(lr){3-5}
Stream & Method & Median & p99 & Mean & Wall (s) & MB \\
\midrule
LmArena & vCache & 7.58 & 14.23 & 5.62 & 1480 & 558 \\
 & \textbf{\sys} & 0.402 & 11.43 & 1.89 & 1323 & 560 \\
\midrule
Classification & vCache & 0.441 & 13.27 & 4.89 & 670 & 504 \\
 & \textbf{\sys} & 0.327 & 10.67 & 0.870 & 1322 & 578 \\
\midrule
SearchQueries & vCache & 0.028 & 13.77 & 3.03 & 6555 & 1158 \\
 & \textbf{\sys} & 0.100 & 6.60 & 1.11 & 6655 & 1245 \\
\bottomrule
\end{tabular*}
\end{table}

\begin{table*}[t]
\caption{Error over $\delta$ of vCache and \sys per cell and the gain of \sys in points over the best baseline that holds $\delta$, with its 95\% CI.}
\label{tab:cells}
\small
\begin{tabular*}{\textwidth}{@{\extracolsep{\fill}}lrrrlrrr}
\toprule
& & \multicolumn{2}{c}{Err./$\delta$} & & & & \\
\cmidrule(lr){3-4}
Stream & $\delta$ (\%) & vCache & \textbf{\sys} & Best baseline & Hit (\%) & Gain & CI \\
\midrule
LmArena & 0.5 & 0.144 & 0.463 & vCache & 8.37 & +13.77 & [12.87, 14.63] \\
 & 1 & 0.122 & 0.520 & vCache & 16.35 & +9.16 & [8.33, 10.15] \\
 & 2 & 0.131 & 0.624 & vCache & 30.49 & +1.94 & [1.18, 2.77] \\
\midrule
Classification & 0.5 & 0.538 & 0.487 & GPTCache 0.97 & 15.95 & +1.52 & [0.86, 2.54] \\
 & 1 & 0.499 & 0.626 & Conformal & 19.21 & +6.28 & [4.38, 8.48] \\
 & 2 & 0.548 & 0.563 & vCache & 33.70 & +0.237 & [$-$0.73, 1.52] \\
\midrule
SearchQueries & 0.5 & 0.426 & 0.596 & Conformal & 2.32 & +1.47 & [0.42, 2.47] \\
 & 1 & 0.412 & 0.660 & Conformal & 5.37 & +1.51 & [0.92, 2.23] \\
 & 2 & 0.387 & 0.694 & Conformal & 10.88 & +0.981 & [0.36, 1.60] \\
\midrule
ClassificationSorted & 0.5 & 0.288 & 0.696 & vCache & 19.57 & +6.14 & [2.16, 9.67] \\
 & 1 & 0.374 & 0.648 & vCache & 38.89 & $-$2.52 & [$-$4.38, $-$1.07] \\
 & 2 & 1.13 & 0.653 & GPTCache 0.97 & 16.26 & +31.45 & [29.55, 32.66] \\
\midrule
Combo & 0.5 & 0.445 & 0.618 & vCache & 3.96 & +1.26 & [0.89, 1.66] \\
 & 1 & 0.410 & 0.639 & vCache & 7.58 & +1.92 & [1.45, 2.51] \\
 & 2 & 0.429 & 0.594 & vCache & 15.53 & +1.71 & [1.16, 2.27] \\
\midrule
Quora QQP & \color{black!55}0.5 & \color{black!55}-- & \color{black!55}-- & \color{black!55}GPTCache 0.97 & \color{black!55}3.49 & \color{black!55}$-$1.32 & \color{black!55}[$-$1.47, $-$1.18] \\
 & 1 & -- & -- & GPTCache 0.97 & 3.49 & +0.411 & [0.26, 0.54] \\
 & 2 & -- & -- & Conformal & 5.59 & +0.667 & [0.26, 1.05] \\
\midrule
Shuffled & all & & & per cell & & +4.10 & [3.57, 4.67] \\
Shuffled & all & & & vCache & & +5.71 & [5.30, 6.02] \\
Drifting & all & & & per cell & & +6.66 & [5.94, 7.28] \\
Shuffled & grid & & & per cell & & +4.32 & [3.91, 4.73] \\
\bottomrule
\end{tabular*}
\end{table*}

\begin{table}[tb]
\caption{Hit rate (\%) and worst error over $\delta$ when Qwen3-8B labels the misses and judges the hits, or only judges the hits on Quora QQP, with the higher hit rate of each pair in bold.}
\label{tab:r2b}
\small
\setlength{\tabcolsep}{2pt}
\begin{tabular*}{\linewidth}{@{\extracolsep{\fill}}llrrrrrr}
\toprule
& & \multicolumn{2}{c}{$\delta=0.5\%$} & \multicolumn{2}{c}{$\delta=1\%$} & \multicolumn{2}{c}{$\delta=2\%$} \\
\cmidrule(lr){3-4}\cmidrule(lr){5-6}\cmidrule(lr){7-8}
Stream & Method & Hit & Err./$\delta$ & Hit & Err./$\delta$ & Hit & Err./$\delta$ \\
\midrule
LmArena, & vCache & 3.37 & 0.533 & 6.70 & 0.574 & \textbf{13.63} & 0.643 \\
order 1 & \textbf{\sys} & \textbf{4.97} & 0.615 & \textbf{8.46} & 0.652 & 12.67 & 0.629 \\
\midrule
LmArena, & vCache & 3.49 & 0.568 & 6.67 & 0.517 & \textbf{13.13} & 0.554 \\
order 2 & \textbf{\sys} & \textbf{4.93} & 0.631 & \textbf{8.84} & 0.680 & 13.03 & 0.616 \\
\midrule
LmArena, & vCache & 3.47 & 0.603 & 6.56 & 0.494 & 13.21 & 0.571 \\
order 3 & \textbf{\sys} & \textbf{4.72} & 0.623 & \textbf{8.52} & 0.680 & \textbf{13.28} & 0.660 \\
\midrule
\multirow{2}{*}{Classification} & vCache & 7.95 & 0.556 & 14.37 & 0.536 & \textbf{28.39} & 0.529 \\
 & \textbf{\sys} & \textbf{13.27} & 0.533 & \textbf{17.34} & 0.481 & 27.10 & 0.550 \\
\midrule
\multirow{2}{*}{SearchQueries} & vCache & 3.02 & 0.370 & 5.36 & 0.385 & 9.82 & 0.393 \\
 & \textbf{\sys} & \textbf{7.38} & 0.555 & \textbf{12.25} & 0.529 & \textbf{19.77} & 0.570 \\
\midrule
Quora QQP, & vCache & 0.229 & 0.194 & 0.407 & 0.194 & 0.800 & 0.202 \\
judge only & \textbf{\sys} & \textbf{2.18} & 0.645 & \textbf{3.91} & 0.598 & \textbf{6.26} & 0.557 \\
\bottomrule
\end{tabular*}
\end{table}

Table~\ref{tab:cells} gives the share of $\delta$ that vCache and \sys spend, as a mean over runs rather than the worst run of Table~\ref{tab:main}, and the gain of \sys over the best baseline that holds $\delta$ in each cell.
Table~\ref{tab:vgain} gives the gains over vCache with other embeddings and budgets, on Quora QQP and under an LRU cap, Table~\ref{tab:mvr} the comparison with MVR-cache and Table~\ref{tab:overhead} the decision cost.
The best baseline of each cell is picked on test.
On Quora at $\delta=0.5\%$ the pick is a fixed threshold whose worst order sits exactly at $\delta$, so its gray row is only an upper bound.

\section{Noise and Scaling}
\label{app:sens}

When a share of the labels each method receives is flipped on the validation split, \sys holds $\delta$ in all nine cells at 5\% and 10\% and keeps more hits than vCache, 7.37\% against 4.91\% and 5.13\% against 4.12\% pooled.
On SearchQueries prefixes of 10,000 to 120,000 queries at $\delta=1\%$, \sys takes 1.01 to 1.09 times the wall-clock time of vCache, and its p99 decision time stays below vCache's.

\section{LLM-Judge Feedback}
\label{app:human}

In the runs of Table~\ref{tab:judge27}, vLLM 0.17 serves Qwen3.8-27B in bf16 with greedy decoding and without thinking on one A800 GPU.
It reads the same prompt and the same truncated answers as Qwen3-8B and judges all 262,362 query pairs of these runs online.
The gains of \sys over vCache are 0.89, 1.82 and 1.81 points at the three budgets, with 95\% CIs of [0.74, 1.04], [1.64, 2.01] and [1.58, 2.01] from a paired bootstrap over queries.
Table~\ref{tab:r2b} gives every budget of the runs in which Qwen3-8B labels the misses and judges the hits, on three LmArena test orders and one order each of Classification and SearchQueries, and of the run on five Quora test orders in which it judges every hit of either cache.
On SearchQueries, whose rows hold no answers, the judge decides whether the new and the cached query share one search intent.

\section{Use of Generative Artificial Intelligence}
\label{app:genai}
Generative artificial intelligence assistants helped draft and edit the text and write the experiment and plotting code.
An image-generation model sketched candidate layouts for the schematic Figures~\ref{fig:teaser}, \ref{fig:motiv} and \ref{fig:overview}, which we redrew as vector diagrams, and drew some of their icons.
The authors checked every statement, number and figure against the experiment records and take full responsibility for the paper.

\clearpage
\end{document}